\documentclass[journal]{IEEEtran}
\usepackage{algorithm}
\usepackage{algorithmic}
\usepackage{amsthm}
\usepackage{subfigure}
\usepackage{epsfig}
\usepackage{graphicx}
\usepackage{url}
\usepackage{multirow}
\usepackage{amssymb}
\usepackage{amsthm}
\usepackage{mathrsfs}
\usepackage{epstopdf}
\usepackage{multicol}
\usepackage{wrapfig}
\usepackage{wrapfig,lipsum,booktabs}
\usepackage{multirow}
\usepackage{amsmath}
\usepackage{amssymb}
\usepackage{xcolor}

\ifCLASSINFOpdf
\else
\fi
\begin{document}
%
\title{When Decentralized  Optimization Meets \\Federated Learning}
%
%
%

\author{Hongchang Gao, 
        My T. Thai,~\IEEEmembership{Fellow,~IEEE,}
        and~Jie~Wu,~\IEEEmembership{Fellow,~IEEE}
\thanks{Hongchang Gao and Jie Wu are with Temple University; My T. Thai is with University of Florida. This submission was accepted to IEEE Network in January 2023.}
}

\maketitle

\begin{abstract}
Federated learning is a new learning paradigm for extracting knowledge from distributed data. Due to its favorable properties in preserving privacy and saving communication costs, it has been extensively studied and widely applied to numerous data analysis applications.  However, most existing federated learning approaches concentrate on the centralized setting, which is vulnerable to a single-point failure.
An alternative strategy for addressing this issue is the decentralized communication topology.  In this article, we systematically investigate the challenges and opportunities  when renovating decentralized optimization for federated learning.   In particular,  we discussed them from the model, data, and communication sides, respectively, which can  deepen our  understanding about decentralized federated learning. 

\end{abstract}


%
\IEEEpeerreviewmaketitle

\section*{Introduction}
%
%
%
%


With the development of Internet-of-Things (IoT) devices and intelligent hardware, various data are generated on these devices every day.  Extracting useful knowledge from these distributed data with machine learning (ML) models to benefit data owners becomes necessary and important. 
Federated learning  (FL) \cite{mcmahan2017communication} provides a feasible way for this distributed ML task with a promise of protecting private information without consuming large communication costs. 
Due to this favorable property, FL has been extensively studied and widely applied to many applications, such as virtual keyboard input suggestion  \cite{yang2018applied} and smart healthcare \cite{xu2021federated}, to name a few. 

In FL, a commonly used  approach to coordinate the collaboration between all participants is  federated averaging (FedAvg). In detail, the central server broadcasts the model parameter to all participants, i.e., data owners.   Each participant updates the received model parameter for multiple iterations by the stochastic gradient computed with its local data, and then uploads the updated model parameter to the central server. After receiving the updated model parameters from all participants, the central server  broadcasts the averaged model parameters to start the next round. With this learning paradigm, all participants can collaboratively learn an ML model without communicating their raw data. As such, the  private information in raw data can be preserved to some extent. Meanwhile, since the model is shared and its size is 
much smaller than the raw data,  the communication cost in FL is reduced significantly.

{Along with such an extensive study of FL, federated optimization was born to further address the computation and communication challenges in FedAvg. Similar to the early phase of FL where focus is on the centralized setting, most of the work in this area concentrates on the parameter-server communication topology, where all participants communicate with the central server. {For instance, \cite{wu2022non} studied the resource and performance optimization in centralized federated learning.} This kind of centralized communication topology, unfortunately, may lead to a single-point failure. In particular, when the number of participants is large, communicating with the central server will cause the communication bottleneck on the central server.} 
With the advance of communication technology, such as 5G/6G, providing fast communication \cite{letaief2021edge} and cloud/edge computation through decentralized computation over IoT and edge devices \cite{lin2019computation}, an alternative strategy is to employ the decentralized communication strategy where all participants perform the peer-to-peer (P2P) communication.  As such, the communication bottleneck will be alleviated. Thus, the decentralized learning paradigm brings new opportunities to the FL development.

\begin{figure*}[h]
	\centering
	\includegraphics[scale=0.45]{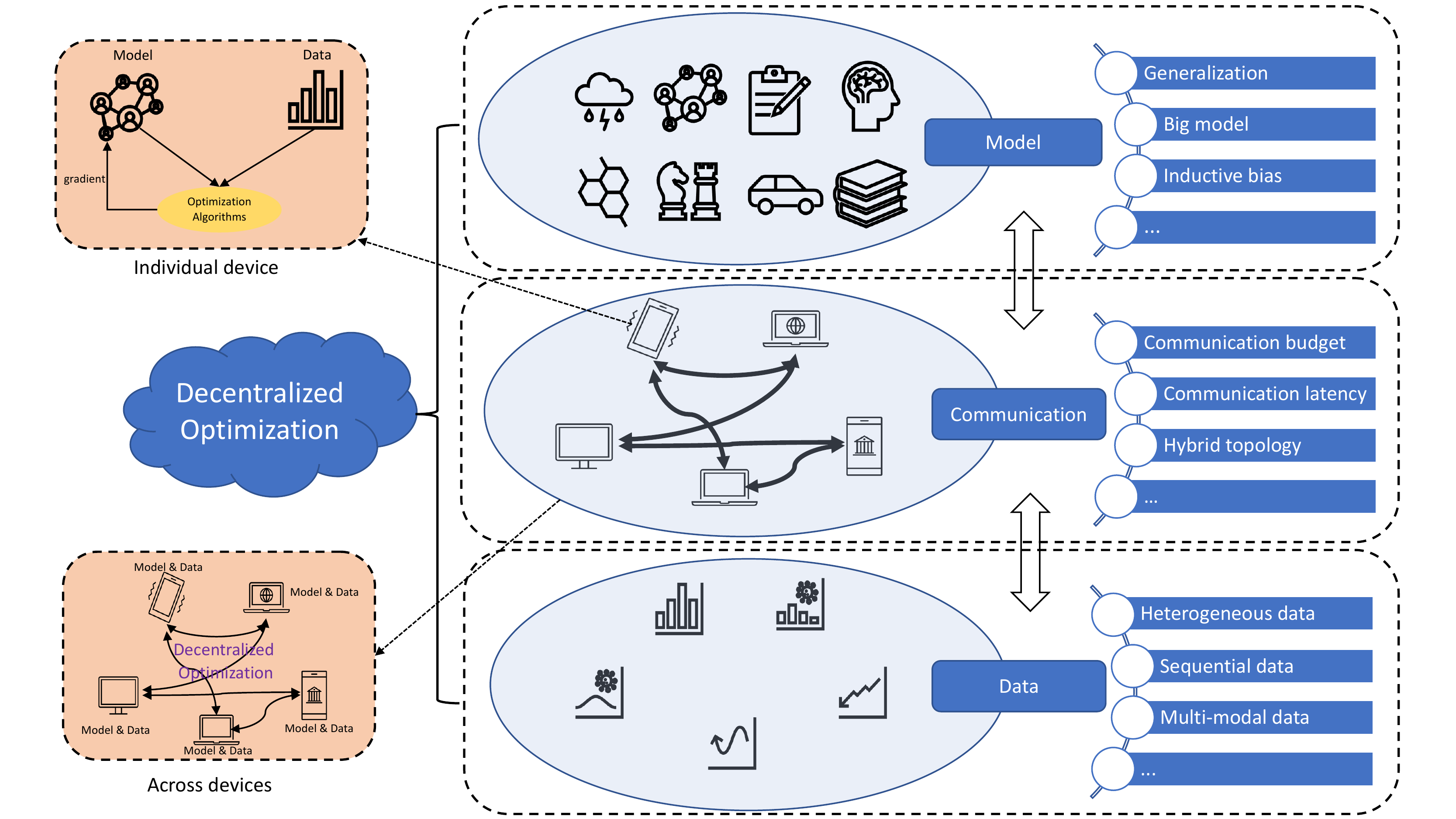}
	\caption{The illustration of decentralized optimization for federated learning. The decentralized optimization unifies FL models, distributed data, and communication together. {In particular, the optimization algorithm on each device bridges the model and data by using the stochastic gradient, which is computed on local data and model, to update local model parameters. Meanwhile, the optimization algorithms across devices can unify all models and data in the entire system via the communication framework to learn a well-generalizing machine learning model. }
} 
	
	\label{overview}
\end{figure*}

In fact, decentralized optimization has been extensively studied in both ML and optimization communities for many years. Numerous decentralized optimization approaches have been developed for the conventional distributed ML model.  However, FL brings new challenges to the conventional decentralized optimization. Just as shown in Figure~\ref{overview}, decentralized optimization serves as the bridge between distributed data and FL models. It should address the unique challenges in the model and data, as well as the issues in itself.
Even though some efforts \cite{gao2020periodic}  have been devoted to facilitating decentralized optimization for FL in the past few years, numerous challenges are still untouched. 

To advance the decentralized FL, in this article, we will review the current development of decentralized federated optimization approaches and then discuss the new opportunities in decentralized FL.  Specifically, this article will focus on the following aspects.
\begin{itemize}
	\item On the model side,  how to improve the FL model's generalization performance with  decentralized  optimization approaches was discussed, pointing out the directions for new algorithmic designs. 
	\item On the communication side, various communication issues when applying decentralized optimization approaches  to FL and  potential techniques for addressing them  were systematically discussed. 
	\item On the data side, we discussed the current challenges  and  future directions when designing new decentralized optimization approaches for FL. 
\end{itemize}
{Following this, we introduce the background of federated learning and decentralized optimization. Then, we discuss the fundamental challenges and potential techniques in  optimization algorithms for decentralized FL. In addition, we introduce challenging issues in communication of decentralized FL. Finally, we discuss how to handle different kinds of data in decentralized FL. }

\section*{Background}

\subsection*{Federated Optimization}
Different from  conventional distributed learning approaches under the data-centre setting, FL faces more computation and communication problems, such as  unstable communication conditions, and highly heterogeneous data distributions, to name a few. A wide variety of federated optimization approaches have been proposed to address these challenging issues in FL.  For instance, to improve the computation complexity, a line of research is to employ advanced gradient estimators, such as the momentum, the variance-reduced gradient \cite{cutkosky2019momentum},  to update the model parameter in each participant. As such, the convergence rate of the improved FedAvg is even able to close to the full gradient descent approach. For instance,  \cite{khanduri2021stem} can achieve the same-order sublinear convergence rate with the full-gradient descent approach for nonconvex problems.  As for the communication complexity, a lot of efforts have been devoted to reducing the communication cost in each communication round and the total number of communication rounds. Moreover, other unique challenges in FL, such as  model personalization, communication security, have also been extensively studied in the past few years.  However, all these approaches just focus on the centralized setting, sharing the  single-point failure issue. 




\subsection*{Decentralized Optimization}
Before the era of FL,  decentralized optimization has already been studied for several decades and has been applied to different domains, such as machine learning, automatic control, etc.  Different from the aforementioned federated optimization approach where a central server coordinates all participants, the decentralized optimization approach does not have such a central server, where each participant directly communicates with its neighboring participants. Based on this communication paradigm, numerous decentralized optimization approaches have been proposed. 

Typically,  according to the specific communication strategy, decentralized optimization approaches can be categorized into two classes. The first category employs the \textit{gossip} communication strategy \cite{lian2017can}. Specifically, each participant computes the gradient based on its local dataset, which is used to update its model parameter.  Then,  each participant  communicates the updated model parameter with its neighboring participants.  The second category employs the \textit{gradient tracking} communication strategy \cite{pu2021distributed}. In particular, each participant introduces an additional variable to track the global gradient, which is employed to update the local model parameter.  As such, in each iteration, the participant should communicate both the model parameter and the tracked gradient. Compared with the gossip-based approach, the tracked gradient  is a better approximation for the global gradient. As such, the gradient tracking is  preferable when the data distribution across  participants is  heterogeneous.

Based on the aforementioned two communication strategies, a wide variety of decentralized optimization approaches have been proposed. For instance, the most straightforward decentralized optimization  approach employs the full gradient to update local model parameters  and then conducts communication at every iteration.  However, the full gradient descent approach suffers from large computational cost in each iteration when  the number of samples is large. To handle the large-scale data, a line of research is to employ the stochastic gradient to update the model parameter in each participant. As such, the computational cost is reduced significantly in each iteration. {In particular, \cite{lian2017can} theoretically demonstrated that decentralized stochastic gradient descent (DSGD) algorithm has almost the same convergence rate with the centralized counterpart for nonconvex optimization problems and the decentralized communication topology only affects the high-order term of the convergence rate of DSGD. Such a favorable convergence rate of DSGD promotes the development of decentralized federated learning in the past few years.} Nevertheless, the stochastic gradient introduces large variance so that the convergence rate is inferior to the full-gradient-based decentralized optimization approach. To address this drawback, multiple variance-reduced approaches have been developed to accelerate the convergence rate of decentralized stochastic gradient descent.

\subsection*{Integration of Decentralized Optimization and Federated Learning}
 Most existing federated optimization approaches concentrate on the centralized setting, which suffers from the intrinsic problems of the centralized system. Thus, integrating decentralized optimization with FL becomes inevitable and promising.  Formally, for decentralized FL, each participant conducts the following steps in each communication round:
 \begin{itemize}
 	\item It computes stochastic gradient based on its local dataset and leverages it to update its model parameter.  This local updates is conducted for $p$ iterations, where $p>1$.
 	\item When the local update is done, each participant communicates its local model parameter or tracked gradient  with its neighboring participants according to the employed communication strategy. 
 \end{itemize}
{Obviously, this decentralized communication strategy avoids communicating with the central server so that there are no communication bottleneck and failure issues in the central server as the centralized federated learning. }
However, this  integration introduces new challenges to decentralized optimization.  In particular, as the bridge between the upper-level FL models  and the lower-level distributed data, decentralized optimization faces with a wide variety of challenges. Just as shown in Figure~\ref{overview},  various FL models require to develop new decentralized optimization approaches to achieve good generalization performance. The complicated data distributions make  conventional decentralized optimization approaches not work. The decentralized communication under the FL setting requires new algorithmic design to handle new communication challenges.  In the following, we will systematically discuss these challenges and potential  techniques to address them from the perspective of the model,  communication, and data, which will help FL researchers and practitioners  deepen their understanding of decentralized FL.

\section*{Decentralized Optimization Meets Models}
The goal of FL is to learn a well-performing ML model for the real-world application.  To deal with different kinds of applications, numerous FL models have been developed. How to ensure the decentralized optimization approach to learn a well-generalized ML model is important and challenging. 
In what follows, we systematically discuss the fundamental challenges  and  potential techniques for addressing them.


\subsection*{How to Achieve Good Generalization Performance?}
The ultimate goal of an ML model is to have good generalization performance. To improve the generalization performance, a lot of efforts have been devoted to the design of ML models. In recent years, the over-parameterized deep neural network has demonstrated superior generalization performance. As such, it has been applied to various FL applications. In turn, it also introduces new challenges to decentralized federated optimization. Specifically, the over-parameterized deep neural network has many local minima where different local minima have different generalization performance. Thus, it is of importance to find the local minima that have good generalization performance.

Existing federated optimization approaches, including both centralized and decentralized ones,  mainly concentrate on the convergence performance. That is how fast an optimization algorithm converges to the local minima. In fact, other than the convergence speed, a decentralized optimization approach  should also have the capability to find the local minima with good generalization performance. Thus, both convergence and generalization performance are of importance when designing  decentralized  optimization approaches for FL. In what follows, we list the essential aspects that need to investigate for the development of decentralized optimization approaches. 

\begin{itemize}
	\item \textbf{Adaptation of existing approaches:}  In recent years, a few new optimization approaches under the single-machine setting have been proposed to pursue the solution that has good generalization performance. For instance, \cite{foret2020sharpness} developed a sharpness-aware optimization approach to find the flat minima, since \cite{keskar2016large} empirically demonstrated that the flat minima enjoy better generalization performance than sharp minima. 
	A straightforward strategy to empower decentralized optimization approaches with the capability of finding well-generalized solutions  is to adapt  existing single-machine approaches, e.g., the sharpness-aware approach,  to the decentralized FL. However, this naive strategy may not work for decentralized FL. For instance,  the sharpness-aware optimization approach has more computational cost due to the maximization and minimization steps in each iteration. Then, the limited computation capability of the participants, e.g., mobile devices, restricts its adaptation to  decentralized FL.  Moreover, the heterogeneous data distribution across participants also introduces new challenges when coordinating the maximization and minimization steps in each iteration. 
	Thus, adapting  existing approaches to decentralized FL requires new efforts to address the computation and communication issues. 
	\item \textbf{Co-design of FL models and decentralized optimization approaches:} Other than adapting existing approaches to decentralized FL, another promising strategy is to co-design the FL model and the decentralized optimization approach to pursue the well-generalized solution. On the one hand, when designing a FL model with good generalization performance,  the model that is easy to be parallelized should be preferable.  Especially, it should avoid employing the global information, e.g., the rank across all samples,  
	since it is difficult for decentralized optimization approaches to get the global information. On the other hand, developing new decentralized optimization approaches for optimizing the well-generalized FL models should be computation-efficient and communication-efficient. For instance, when the FL model requires the global information, it is necessary to employ some  strategies to approximate it to  avoid the frequent communication across  participants. 
\end{itemize}

\subsection*{How to Handle Big Models?}
To pursue the well-generalized ML model, a surge of interest focuses on developing big models. Specifically, the big model has a huge number of model parameters and it is trained with the huge volume of training data. For instance, GPT-3 \cite{brown2020language} has 175 billion model parameters and it is trained with 45TB training data. 
With such large model size and training data, big models enjoy superior generalization performance.  For instance, GPT-3 can achieve great generalization performance for few-shot and zero-shot learning. Thus, adapting big models for FL can benefit a wide variety of real-world applications. 

However, the big model incurs new challenges for decentralized FL due to its large model size and the huge volume of training data. Directly deploying big models to decentralized FL seems infeasible since the computation capability of the participants  is limited. Moreover, training big models with decentralized FL requires a huge number of participants to get enough training data. Such kinds of large-scale distributed data is more likely  to be heterogeneous. Without a central server, it is difficult for a decentralized FL system to get the global information to address the heterogeneous issue. Thus, it is difficult to train big models with decentralized FL. 

How to apply the promising big model to decentralized FL requires new efforts in the design of learning paradigms and  corresponding decentralized optimization approaches. 
In what follows, we discuss several prominent aspects to address this unique challenge.
\begin{itemize}
	\item \textbf{Zeroth-order approaches:}  Since it is infeasible to train a big model under the decentralized FL setting due to the huge model and data size, a potential strategy for leveraging big models is to employ the pre-trained big models as a service provider. In particular,  rather than training a big model from scratch, we can directly utilize the pre-trained big model to benefit the small model training. For instance, as shown in \cite{sun2022black}, the big language model GPT-3 can generate an augmented sample for the input sample, which can be utilized for prompt tuning. Therefore, we can put the big model on each participant. Then, the participant can  leverage the model output from the local input data to optimize the parameter of the prompt learning part. Since the model parameters of big models are typically not accessible,  we need to develop the zeroth-order decentralized optimization approach for this kind of task.  Currently, there are very few works about zeroth-order decentralized optimization approaches for FL. Thus, the systematic investigation about the computation and communication complexities of zeroth-order approaches is of immense importance and necessity. 
	\item \textbf{Low-dimensional approaches:} Since the big model has a large number of model parameters, a potential strategy to train or fine-tune this kind of big models is to optimize model parameters in the low-dimensional space. For instance, one can employ the sketching method to project model parameters in a low-dimensional subspace and then perform optimization in such a subspace. However, this strategy causes new challenges for decentralized FL.  
	For instance, how fast the low-dimensional approach will converge is not clear. Hence, new efforts should be devoted to the systematic investigation on the algorithmic design and theoretical analysis about the low-dimensional decentralized federated optimization approaches. 

\end{itemize}

\subsection*{How to Deal with Inductive Biases? }
Although the big model is promising in improving generalization performance, it requires powerful computation capability and a large volume of training data. For some real-world FL applications, it is difficult to obtain large-scale training data. For instance, the healthcare data is typically not large enough to train a big model.  To address this issue, an alternative strategy is to incorporate the inductive bias to  regularize the model to have the desired performance.  
Specifically, an important kind of inductive bias is to make a FL model to capture the intrinsic structure in the data. For instance, the convolutional neural network should be invariant to the translation and rotation of input samples.  A high-dimensional model should be aware of the low-dimensional subspace. 
Moreover, another important inductive bias is to make a FL model to capture the domain knowledge in specific applications. For instance, the graph neural network for molecular graphs should be aware of the valid subgraph structure.  

To incorporate inductive biases into FL models, some models use  constraint to deal with them. For instance, the low-rank matrix completion model has a trace-norm constraint to pursue a low-rank solution. Most existing decentralized optimization approaches concentrate on the unconstrained problem. How to solve the constraint problem under the decentralized and periodical communication condition is still under explored, which requires systematic investigation as follows.
\begin{itemize}
	\item \textbf{Convex constraint:}  The convex constraint is widely used in ML models to deal with inductive biases, such as the low-rank constraint. To solve the FL model with convex constraint, a critical challenge is the computation complexity when dealing with the constraint. The possible strategy includes the projection gradient descent and conditional gradient descent. However, how these approaches converge under the decentralized FL setting is still unclear. Thus, it is necessary to adapt those algorithms to decentralized FL and investigate their computation and communication complexities.
	
	\item \textbf{Non-convex constraint:} Compared with the convex constraint, non-convex constraint is much more difficult to solve since the convex combination of the solutions may not satisfy the constraint. Thus, it requires new algorithmic design to deal with the non-convex constraint. Especially, it would be better if the new algorithm does not require the global information since it is difficult to get it under the decentralized FL setting. Moreover, more efforts should be devoted to the investigation of the computation and communication complexities of this kind of decentralized federated optimization approaches. 
\end{itemize}

\subsection*{More Challenges}
Other than the generalization issue, there are some other challenges in decentralized FL, e.g., the fairness issue. In particular, even though machine learning  has achieved remarkable success in many real-world applications, it has been observed that the prediction result could have discrimination for minority groups. To address this issue, new decentralized optimization algorithms should be explored to learn a fair machine learning model. More specifically, some efforts have been made to developing new machine learning models, which are able to guarantee individual and group fairness. Those fair machine learning models cause new challenges for decentralized optimization. For instance, some of those new models belong to the min-max optimization problem, rather than the traditional minimization problem. How to facilitate them to decentralized FL is under-explored. Especially, how the communication period affects the convergence rate is still unclear. Therefore, more endeavor is needed to establish the foundations of decentralized federated optimization for these emerging machine learning models. 

Moreover, in decentralized FL,  each participant might optimize multiple tasks simultaneously, i.e., the multi-objective optimization problem. How to solve the multi-objective optimization problem under the decentralized FL setting is still unexplored. Especially, the intrinsic properties in decentralized FL bring unique challenges. For instance,  different participants pay different attention to those objectives. How to differentiate the tasks should be considered when designing new decentralized  optimization approaches for this kind of FL applications.  Meanwhile, different tasks might have different inductive biases. How to deal with those inductive biases simultaneously should also be investigated under the decentralized FL setting.  

\section*{Decentralized Optimization Meets Communication}
In FL, different participants have different communication conditions, such as limited communication budget, large communication latency, to name a few. Adapting decentralized optimization approaches to these complicated communication conditions is of importance and necessity.

\subsection*{Limited Communication Budget}
For decentralized optimization, each participant should communicate its local model parameters or gradients with its neighboring participants. When the size of FL models is large, the communication cost will be high, which can degenerate the empirical convergence speed.  Thus, a  core research question is to reduce the communication complexity. In fact, numerous efforts have been made to improve the communication complexity of the centralized FL. However, they are not applicable to the decentralized setting, especially how those techniques affect the convergence rate of decentralized optimization approaches is not clear. To address the communication complexity issues, the following aspects should be investigated. 
\begin{itemize}
	\item \textbf{Reducing communication rounds:} To improve the communication complexity, a promising strategy is to reduce the number of communication rounds. However, the periodic communication incurs new challenges for decentralized optimization with the gradient tracking technique. In particular,   the tracked gradient in conventional decentralized optimization approaches is computed based on the local gradients in two consecutive iterations. With the periodic communication, it is unclear whether the gradients in two consecutive iterations or communication rounds should be used. Thus, it is necessary to investigate different algorithmic designs and how they affect the converge rate and communication complexity. 
	\item  \textbf{Reducing communication cost:} Another commonly employed strategy is to compress the communicated variables. As such, the communication cost in each communication round is reduced significantly. How to apply the compression techniques to the decentralized communication approach in the presence of periodic communication is still under-explored. Thus, it is promising to investigate how to combine the compression technique and  periodic communication strategy to reduce the communication complexity of decentralized optimization approaches. 
\end{itemize}

\subsection*{Large Communication Latency}
Since different participants possess different computation and communication capabilities, there usually exists large communication latency in a FL system, which can slow down the empirical convergence speed of decentralized optimization approaches. Even though some methods have been proposed for the centralized FL, they are not applicable to the decentralized FL due to the decentralized communication strategy. Thus, it is necessary to develop new decentralized optimization approaches to deal with the large latency issue in FL. 

A promising direction to address this challenge is the asynchronous communication strategy, where each participant overlaps its computation and communication. As such, the empirical convergence speed will be improved. However, there exist new challenges when employing the asynchronous communication strategy for decentralized optimization. Especially when employing the gradient tracking technique, both model parameters and tracked gradients should be communicated. As such, there exists asynchrony between computation and communication, as well as two communication procedures for model parameters and gradients. Thus, it is challenging and important to investigate how the asynchronous decentralized optimization approaches for FL converge and how large communication latency it can admit. 

\subsection*{Hybrid Communication Topologies}
In some real-world FL applications, both the centralized and decentralized communication topologies are utilized to leverage their advantages. In particular, the decentralized communication in a P2P structure can alleviate the single-point failure issue in the centralized one. In turn, the centralized communication is able to benefit the convergence speed. Thus, it is necessary to develop new federated optimization approaches for the hybrid  communication topology. Key issues in topology design are to decide the number of communicating neighbors and to choose these neighbors.

On the one hand, under the FL setting, the new decentralized optimization approach for improving the generalization performance should be developed, and its convergence rate requires to study. In particular, how the spectral gap affects the convergence rate needs to investigate.  On the other hand, the communication-efficient decentralized optimization approach under the FL setting should be studied, and the convergence rate should be established.

\section*{Decentralized Optimization Meets Data}

In FL, the training data is much more complicated than the data-centre setting. For instance, the data might be highly heterogeneous across all participants. 
  Moreover, in some applications, such as autonomous driving, the data are sequentially generated. All these scenarios bring new challenges for decentralized federated optimization.

\subsection*{Heterogeneous Data}
When different participants have different data distributions, the stochastic gradient at each participant is significantly different from the global gradient. As such, the local model parameters at different participants will converge to different stationary points. Thus, it is of importance to alleviate the heterogeneous data distribution issue to guarantee  convergence.  However, there does not exist a central server to get the global information. Thus, alleviating the heterogeneous issue for decentralized FL requires new algorithmic designs.

	In traditional decentralized optimization, a commonly used approach to address the aforementioned issue is the gradient tracking technique. In particular, 
	 the gradient tracking technique requires to communicate both model parameters and gradients. As such, the gradient at each participant is able to track the global gradient. The effect from the heterogeneous data distribution can be alleviated to some extent.  However, under the FL setting, the communication is performed  periodically. Thus, whether the gradient tracking technique can effectively track the global gradient is unclear. Therefore, it is necessary to investigate how  these two strategies affect the heterogeneity term in the convergence rate.  Moreover, unlike the centralized FL where it is easy to obtain the global information to alleviate the heterogeneous issue, new strategies, such as combining centralized and decentralized communication, should be investigated to address this issue.

\subsection*{Sequential Data}
Most existing decentralized FL models concentrate on the independent data, where different samples are independent of each other. However, in some real-world applications, there exists dependence between different samples. For instance, in autonomous driving, the car interacts with the environment, and then the data is sequentially generated. As such, there exists dependence among this kind of sequential data. In fact, this kind of application belongs to multi-agent reinforcement learning  when there are multiple self-driving cars. 
Typically, since the self-driving car  need to interact with its surrounding cars, it is  appropriate to formulate this application as a decentralized FL task. 

Traditional decentralized optimization approaches for FL just focus on the standard gradient, ignoring the dependence in the data. Thus, it is necessary to develop new decentralized optimization approaches for the federated sequential decision task. In particular, a potential direction is to study the decentralized stochastic gradient descent (SGD) with periodic communication for Markov process. Specifically, in the sequential decision task, it is typically assumed that the decision procedure follows the Markov process. As such, we should investigate how the decentralized Markov SGD converges under the periodic communication strategy. Moreover, the communication complexity should also be investigated. In particular, how the communication period affects the convergence rate should be investigated to benefit the FL practitioners.

\subsection*{Multi-modal Data}
The multi-modal data is very common in real-world FL applications.  Different modalities might be distributed in different participants. For instance, the healthcare data could include different types of diagnosis records, and these records sometimes are distributed in different hospitals since  a patient may take CT scan in one hospital and get diagnosis in another hospital.  To make predictions for these kinds of patients, we need to unify the features from all hospitals. To address such kinds of multi-modal data, the centralized FL developed  the vertical FL paradigm to coordinate the feature learning across all participants. However,  under the decentralized FL setting, the data owners of different modalities might not be connected directly, and there is no central server to coordinate the collaboration among data owners. Thus, the existing vertical FL paradigm does not work for the decentralized setting. New learning paradigms should be investigated to address the dependence among different data owners when making predictions. { A potential solution to address this inter-device dependence issue is to employ the hybrid communication topology where the global communication is conducted but infrequently. As such, each local device could leverage the outdated multi-modal data to do prediction.} Correspondingly, new decentralized optimization approaches for this kind of FL application should be developed to address the dependence among multi-modal data, e.g., how large the outdated period can be admitted without hampering the convergence rate.

\section*{Conclusions}
In this article, we provide a comprehensive discussion about decentralized optimization approaches for federated learning. In particular, the integration of decentralized optimization and federated learning brings new challenges and opportunities.  The decentralized optimization is able to address the intrinsic problems of the conventional federated learning system. In turn, federated learning provides new opportunities to boost the development of decentralized optimization. We systematically investigate these challenges and opportunities from different perspectives, including the model, data, and communication, which points out the potential research directions and can help readers deepen their understanding about decentralized federated learning.

\section*{Acknowledgement}
This research was supported in part by NSF grants CNS 2214940, CPS 2128378, CNS 2107014, CNS 2150152, CNS 1824440, CNS 1828363, CNS 1935923, CNS 2140477, and IIS 939725.

\bibliographystyle{plain}
\bibliography{IEEEabrv}

\section*{Biographies}
\noindent Hongchang Gao (hongchang.gao@temple.edu) is an assistant professor in the department of computer and information sciences at Temple University, USA. He received his Ph.D. degree in electrical and computer engineering from University of Pittsburgh in 2020.   His research interests include machine learning, deep learning, and optimization. 


\noindent My T. Thai [IEEE Fellow] (mythai@cise.ufl.edu)
is the UF Foundation Professor and Associate Director of Nelms Institute for the Connected World at the University of Florida. Her research interests include trustworthy AI, quantum computing, and optimization.

\noindent Jie Wu [IEEE Fellow] (jiewu@temple.edu) is the Laura H. Carnell  Professor and Director of  Center for Networked Computing at Temple University. He is  a member of the Academia Europaea. His research interests include mobile computing and wireless networks, network trust and security, applied machine learning, and cloud computing.

\end{document}